# Federated Neuroevolution O-RAN: Enhancing the Robustness of Deep Reinforcement Learning xApps

Mohammadreza Kouchaki, Aly Sabri Abdalla, Vuk Marojevic
Electrical and Computer Engineering, Mississippi State University, USA
mk1682@msstate.edu, asa298@msstate.edu, vm602@msstate.edu

*Abstract*—The open radio access network (O-RAN) architecture introduces RAN intelligent controllers (RICs) to facilitate the management and optimization of the disaggregated RAN. Reinforcement learning (RL) and its advanced form, deep RL (DRL), are increasingly employed for designing intelligent controllers, or xApps, to be deployed in the near-real time (near-RT) RIC. These models often encounter local optima, which raise concerns about their reliability for RAN intelligent control. We therefore introduce Federated O-RAN enabled Neuroevolution (NE)-enhanced DRL (F-ONRL) that deploys an NE-based optimizer xApp in parallel to the RAN controller xApps. This NE-DRL xApp framework enables effective exploration and exploitation in the near-RT RIC without disrupting RAN operations. We implement the NE xApp along with a DRL xApp and deploy them on Open AI Cellular (OAIC) platform and present numerical results that demonstrate the improved robustness of xApps while effectively balancing the additional computational load.

## I. INTRODUCTION

In the face of today's dynamic digital landscape characterized by increasing interconnectivity and technology convergence, traditional rigid and closed network deployments no longer suffice to meet the demands of modern communication systems. To address these challenges, open radio access network (O-RAN) offers a groundbreaking approach to mobile networks, aiming to revolutionize the industry by enhancing flexibility, interoperability, and cost efficiency. Unlike traditional networks that heavily rely on proprietary and tightly integrated solutions, O-RAN introduces a novel disaggregated RAN architecture with open interfaces, effectively decoupling the hardware and software components involved. This decoupling enables greater agility and adaptability, allowing easier integration of new functionalities and technologies.

O-RAN also introduces RAN Intelligent Controllers (RICs), two logical control units enabling monitoring and optimizing RAN configurations at non-real time (RT)—seconds to minutes or more—and near-RT—10 ms to 1 second—time scales, respectively [1]. These intelligent controllers serve as the orchestrators of the xApps and rApps microservices embracing artificial intelligence (AI) and machine learning algorithms. rApps are deployed in the non-RT RIC which is part of the service management and orchestration asset of an O-RAN operator. They enable long-term network optimization, policy-based management, and data-driven orchestration. xApps, on the other hand, live in the near-RT RIC which interfaces with the disaggregated RAN through O-RAN's E2 interface. They may support near-RT decision making and optimization that influence short-term RAN operations, such as resource allocation, interference monitoring, and mobility control.

While rApp type network orchestration is common in cellular networks, this paper emphasizes on improving the reinforcement learning (RL) and deep RL (DRL) performance for xApps. RL and DRL, in particular, empower the network to learn, adapt, and optimize its operations. The incorporation of deep neural networks (DNNs) as function approximations that are trained through gradient-descent methods has been widely adopted for DRL algorithms for solving complex optimization problems. For instance, RL and Federated RL models leveraging the actor-critic approach have been effectively employed for resource management [2] [3], an RL-based cell association xApp [4], and a traffic steering xApp [5] in O-RAN.

One known difficulty of RL algorithms is the temporally disjointed nature of its reward feedback system. Rewards are often sparse and substantially delayed, leading to the well-known credit assignment problem [6]. This issue is particularly challenging for gradient-based techniques, as they inherently assume immediate and frequent feedback for efficient learning. Striking a balance between exploration and exploitation is another core challenge of RL. Gradient-based techniques often fall into local optima, focusing on known rewards while inadequately exploring the environment for potentially higher returns [6]. With each agent policy update, the data distribution continuously changes, which is in contrast to the static data distribution assumed by gradient-based learning, leading to potential instability and reduced performance. Moreover, RL lacks labeled samples that are prevalent in supervised learning, relying instead on scalar reward signals [7]. These challenges form significant drawbacks for RL, potentially undermining their efficacy for network and resource management in O-RAN systems [5]. The above problems apply to DRL, which we consider in this paper because of its superior in exploring the large action spaces of advanced wireless network resources.

The integration of evolutionary algorithms (EA) within DRL models offers key advantages, including the ability to handle sparse and delayed rewards through a holistic evaluation as well as to promote global exploration for avoiding local optima [8]. EA-based techniques moreover provide stable learning in non-stationary environments and leverage gradient-free optimization, addressing many limitations of traditional gradient-based approaches. EA is a bio-inspired metaheuristic optimization model that replicates the process of natural evolution. It is characterized by its principal operations: mutation, reproduction, selection, and recombination. These operations when applied to DRL guide the DNN weight adjustments, steering the AI model toward optimal performance.

Efforts in EA have led to substantial advancements in DNNs. There have been few initial studies that provide insights into how EA can be applied to O-RAN. For instance, the work presented in [9] highlights some of the design choices for combining RL with EA to improve network efficiency and adaptability by simplifying the inference models for dynamic RAN optimization. The authors of [10] propose an EA-based DRL framework that optimizes slice management while avoiding service level agreement violations. These works are theoretical in nature, and lack testbed demonstrations.

The performance of DRL-based xApps can be enhanced by introducing a novel hybrid EA to evolve the DNN architecture, a process known as neuroevolution (NE) [8]. This empowers xApps to better navigate the complexities of learning in unknown environments, thereby improving the efficiency and robustness of O-RAN. However, the increased complexity resulting from this integration needs to be addressed. Therefore, in this paper, we introduced a federated NE enhanced DRL based solution for O-RAN's near-RT RIC, leveraging the O-RAN architecture and its Kubernetes (K8s) container deployments to enhance the computing efficiency and robustness of DRL based xApps. The proposed Federated O-RAN enabled NE-enhanced DRL (F-ONRL) entails a single EA optimizer xApp, which utilizes parallel computing resources within the near-RT RIC. This enables the parallel integration of the optimizer xApp with other xApps that are dedicated to specific RAN control functions. We deploy this framework on a real-world O-RAN testbed, enabled by the Open AI Cellular (OAIC) platform, to validate the proposed F-ONRL solution.

The rest of the paper follows as such: Section II presents the proposed F-ONRL xApp framework. Section III details the experimental deployment of the proposed framework and obtained results. Section IV introduces the challenges and future directions and Section V offers concluding remarks.

## II. Proposed F-ONRL xApp Framework

NE employs EA to generate parameters or learning rules and guide the DRL model toward more effective strategies. The two main categories of NE are direct and indirect encoding. Direct encoding explicitly represents every element of the DNN architecture, neurons, connections, and their weights in the genome. This means each gene directly influences a specific aspect of the network. This approach offers precise control over the DNN architecture and its parameters but can be computationally expensive for complex state-action spaces. Instead of directly representing the DNN architecture, indirect encoding uses a more compact genome that encodes a set of rules or a blueprint for constructing the network architecture.

We opt for direct encoding for the xApp optimization framework because it is well-suited for the scalability demands of O-RAN. This approach allows for more targeted adjustments, ensuring that the DRL model explores a broad solution space while maintaining the high performance necessary for near-RT decision-making and continuous learning in large-scale wireless networks. We employ the genetic algorithm (GA), a form of EA known for its robustness in traversing complex search spaces, for fostering diversity in the solution exploration and mitigating premature convergence to sub-optimal solutions. However, the NE approach suffers from two major limitations including slow learning rates and high computational demands [11]. We therefore introduce the F-ONRL employing separate computing engines within the K8s platform used for deploying O-RAN's near RT RIC. This setup supports concurrent NE and DRL operations.

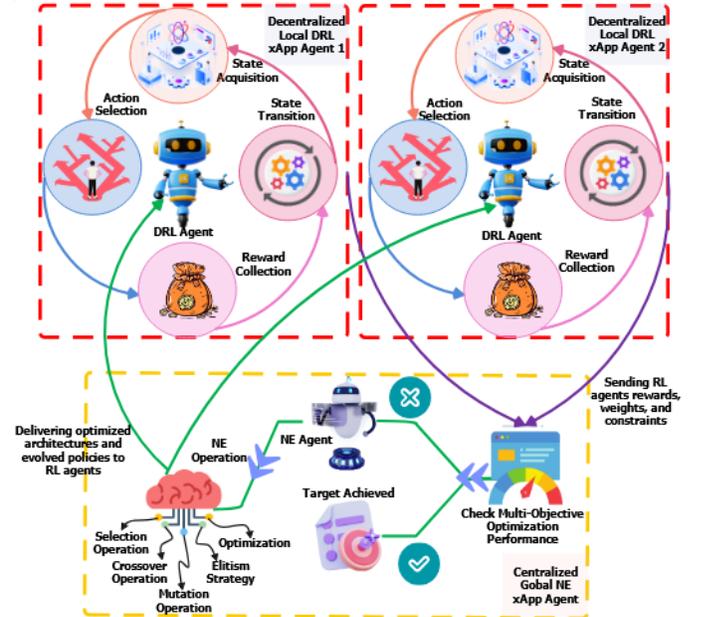

Fig. 1: The DRL and NE agent processes and interactions, showing two distinct DRL agents being controlled by a single NE agent.

### A. Design, Deployment, and Operation

F-ONRL uses the global search capabilities of NE while leveraging the local optimization efficiency of DNNs. The NE and DRL agents are encapsulated as separate, containerized xApps with two main functionalities: The NE xApp accelerates the DNN training/weight adaptation process while the DRL xApps perform primary RAN control tasks, such as load balancing, slicing, or resource optimization. The DRL agent xApps therefore continuously interact with the O-RAN environment. They perform state acquisition, action selection, action execution, reward collection, and state transition for their respective RAN control tasks. Each DRL xApp thus maintains a distinct Markov Decision Process with task-specific states, actions, and reward structures tailored to its unique optimization objective. It monitors the reward over specified intervals and sends its AI model parameters to the NE xApp whenever its performance drops. The NE xApp then refines the DRL parameters and sends the updated ones back to the DRL xApp for immediate deployment, without interfering with the RAN control operations. More precisely, each DRL agent independently assesses its own performance and upon detecting insufficient improvement or persistent under performance relative to set targets, asynchronously triggers the centralized NE xApp for evolutionary optimization. The optimized network parameters are returned to the DRL xApp. Fig. 1 illustrates this.

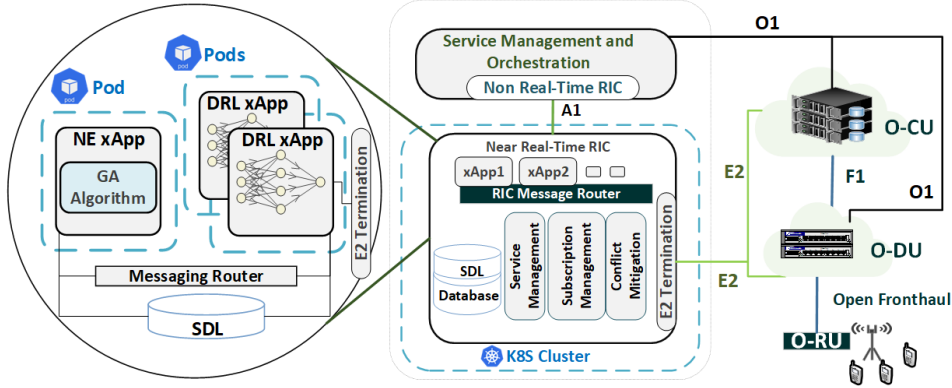

Fig. 2: F-ONRL deployed in the near-RT RIC of the O-RAN architecture

In the current version of our solution, the NE xApp sequentially handles optimization requests; thus, the optimization of neural network parameters for each DRL agent is performed individually based on the received DNN parameters, reward information, and related performance metrics.

The interactions between the RAN controller xApps and the NE xApp follow federated learning (FL) principles, balancing exploration and exploitation [3], The DRL xApps are decentralized actors that continuously collect rewards and execute RAN control actions. The NE xApp is a centralized global optimizer that asynchronously optimizes the DLR model parameters of a single or multiple DRL agents. This asynchronous workflow ensures that DRL actions remain unaffected during parameter adaptation, maintaining responsiveness and stability.

F-ONRL leverages the K8s platform's containerization and a custom resource definition (CRD) to allocate dedicated resources to each DRL and NE agent. This enables encapsulating DRL and NE xApps in separate K8s pods for their independent deployment, operation, and control of dedicated resources within the K8s cluster (Fig. 2).

### B. Detailed Workflow

The F-ONRL process begins with the DRL agent performing its core operations: state acquisition, action selection, action execution, reward collection, and state transition. This loop operates continuously, allowing the DRL agent to interact with its environment to fulfill its designated tasks and iteratively collects feedback from its actions (Fig. 3).

Each DRL agent evaluates its performance based on these feedback and predefined performance metrics, which are typically represented as reward function targets. The average return of the reward function, calculated over the state-action-reward cycles within an NE interval, serves as a key indicator of the agent's effectiveness. If the agent continues to exploit its current knowledge (captured by the DNN configuration and weights) without achieving sufficient improvement in the average return, it signals the need for further exploration. The NE agent facilitates this exploration by introducing new configurations of DRL model parameters, potentially uncovering more effective solutions for improving the DRL agent's decision making (RAN control) performance. It optimizes the DRL agent's network configuration based on the performance feedback from the DRL agent and sends the optimized parameters back to the DRL agent. The process continues until the system achieves the desired performance level or until reaching a cut-off point. A checkpoint at the end of the GA loop, as shown in Fig. 3, evaluates performance to decide whether to terminate the NE evolution or continue refining the parameters.

The workflow of the NE agent is centered around three main GA operations—selection, crossover, and mutation—that are supplemented by an elitism strategy. The selection operation drives the optimal search in the DNN weight space by prioritizing the fittest individuals for reproduction based on their evaluated fitness. The crossover operation, modeled after genetic recombination, generates an offspring by averaging weights from two-parent solutions. This exploration of intermediate solutions enriches the search process by broadening the population's genetic diversity. Mutation introduces random alterations to the offspring weights, enabling further exploration of the weight space and avoiding local optima. The elitism strategy ensures that top-performing individuals are carried over across generations for population fitness improvement. Through iterative evolution over a defined generation, these operations produce a population of solutions with improved DRL model parameters.

F-ONRL integrates two synergistic mechanisms for parameter tuning and resource management, enabling the NE agent to dynamically adjust its computational effort based on the DRL agent's performance and system resource constraints.

**Performance-Based Adaptation with Iterative Refinement:** The DRL agent computes the difference between the desired performance metric, which is defined as the system's reward, and the current situation as an indication metric. This metric guides the NE agent to select one of three predefined GA parameter sets (low, medium, high) that correspond to increasing levels of computational effort (population size, mutation rate). Low-tier parameters are applied for minimal adjustments, while higher tiers address significant reward stagnating or divergence. This dynamic adjustment ensures that the NE agent dedicates more resources to exploration only when necessary to balance performance and complexity.

**Dynamic Resource Allocation:** At the beginning of each NE refinement, the NE agent considers the GA parameter set indication (high, medium, low) based on the DRL agent's reward and queries the K8s API and CRD to gain knowledge of resource requests, limits, and the current near-RT RIC's K8s cluster utilization. The obtained resource information assists the NE agent in adjusting key GA parameters (population size, mutation rate, and number of generations) via a *scaling factor*. This approach ensures that GA operations are resource-aware and balanced against system constraints.

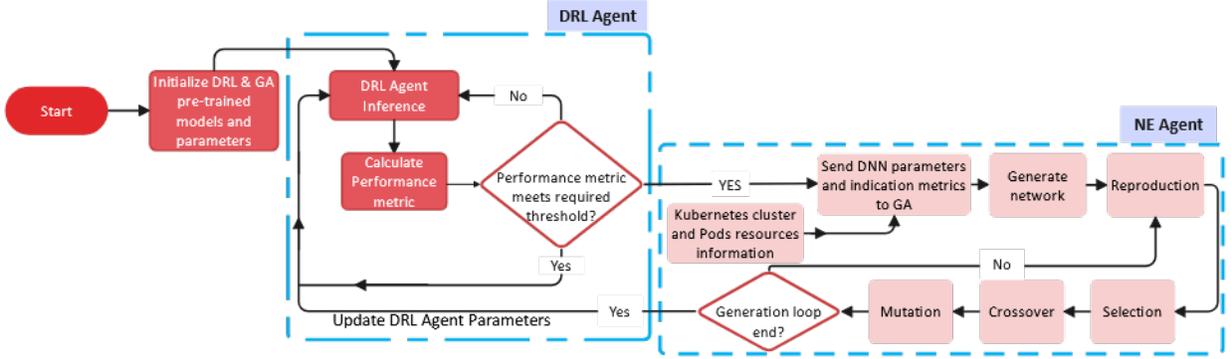

Fig. 3: F-ONRL workflow.

## C. Data Flow

O-RAN defines the components, processes, and interfaces for the non and near-RT RICs to interact with the disaggregated RAN, which consists of the O-RAN or open central unit (O-CU), distributed unit (O-DU), and radio unit (O-RU), as illustrated in Fig. 2. The O-CU is responsible for higher-layer functions such as radio resource control and handover management. It interfaces with the O-DU via the F1 interface. The O-DU handles real-time processing of lower protocol layers, the higher part of the physical layer and the medium access control layer, and interfaces with the O-RU for signal processing and baseband operations of the lower part of the PHY. The O-RU manages the radio frequency front-end, including antenna control and signal transmission/reception, and connects to the O-DU via the open fronthaul interface [12].

Data and control messages between the near-RT RIC and the E2 nodes (O-CUs and O-DUs) are exchanged employing the E2 protocol over the E2 interface. O-RAN's E2 interface thus facilitates the exchange of RAN data and DRL actions. These interactions between the RAN and the xApps leverage the shared data layer, which is supported by a time-series database (InfluxDB), and the RIC message router (RMR). The communication between the DRL and NE agents is also supported by the RMR, which uses the K8s networking model to ensure timely DNN weight evolution. K8s inherently supports load balancing and resource auto-scaling for efficient resource utilization by the DRL and NE agents while maintaining reliable performance under varying workloads.

Deploying the NE optimization as an xApp within the near-RT RIC ensures timely parameter evolution through secure internal message exchanges via the RMR protocol, mitigating security vulnerabilities associated with external communications. Conversely, an rApp implementation of the NE agent would incur additional latency and expose the parameters to potential security vulnerabilities. Future F-ONRL implementations can incorporate fitness-gated parameter deployment to filter unstable NE outputs, RMR-level hash-based verification for advanced NE–DRL message exchange security, and checkpoint-triggered fallbacks based on reward divergence.

## III. EXPERIMENTAL DEPLOYMENT AND RESULTS

### A. Experimental Platform

OAIC provides a community research platform for developing a variety of cellular traffic patterns, channel conditions, and user behaviors experiments. OAIC is based on open-source 5G RAN software and the O-RAN Alliance specifications to facilitate rapid prototyping and testing of new AI-RAN solutions, enabling 6G research [13]. It implements the standard near-RT RIC based on O-RAN Software Community's software, E-Release and the E2 interface. Data and control message exchanges between the near-RT RIC and the E2 nodes follow the E2 application protocol.

The srsRAN software Version 21.10 runs on Ubuntu 20.04 LTS with Docker Engine Version 24.0 on a commercial off-the-shelf computer. We use an Intel Xeon server, 256 GB DDR5 RAM, and a dual NVIDIA RTX A6000 GPU. Instead of SDR hardware, we employ Zero Message Queue (ZMQ) for message passing between the srsrRAN and the srsUE software.

The testbed is configured with varying network parameters, including neighboring base stations and traffic intensity. Key parameters, such as channel quality indicator (CQI), data rates, and user distribution, are modeled based on standard 5G benchmarks to ensure consistency with real-world deployments. This setup facilitates creating dynamic network conditions for evaluating different traffic management strategies.

### B. RAN Control xApp

We validate F-ONRL using single and multiple E2 nodes managed by xApps in the near-RT RIC. The xApp under test is designed to optimize RAN operations by performing traffic steering and dynamic resource allocation. In each cycle, it collects metrics from E2 nodes, including CQI, data rates, and the number of active and connected UEs. These metrics represent channel conditions, user throughput, and network load distribution, providing the xApp with the necessary information to make informed decisions in near-RT time.

TABLE I: Experiment configurations.

| Network Parameters | Experiment1 | Experiment2 | Experiment3 | Experiment4 | Experiment5 | Experiment6 | Experiment7 |
|---|---|---|---|---|---|---|---|
| Model Name | NE-DQN(1st) | NE-DQN(2nd) | NE-DQN(3rd) | NE-A2C(1st) | NE-A2C(2nd) | NE-A2C(3rd) | NE-MARL |
| NE Interval | 125 | 125 | 125 | 125 | 125 | 125 | 125 |
| Indication Metric | Low | Medium | High | Low | Medium | High | High |
| Scaling Factor | 1 | 1 | 1 | 1 | 1 | 1 | 1 |
| Generation Number | 50 | 100 | 300 | 50 | 100 | 300 | 500 |
| Population Size | 40 | 70 | 125 | 40 | 70 | 125 | 125x2 |
| Elitism Number | 1 | 2 | 5 | 1 | 2 | 5 | 5 |
| GA Mutation Rate | 0.01 | 0.1 | 0.2 | 0.01 | 0.1 | 0.2 | 0.2 |
| GA Crossover Rate | 0.3 | 0.5 | 0.8 | 0.3 | 0.5 | 0.8 | 0.8 |
| Learning Rate | 0.005 | 0.005 | 0.005 | 0.001 | 0.005 | 0.001 | 0.001 |
| Gamma ($\gamma$) | 0.95 | 0.95 | 0.95 | 0.99 | 0.99 | 0.99 | 0.99 |
| Max Downlink Traffic per UE | 1 Mbps | 1 Mbps | 1 Mbps | 1 Mbps | 1 Mbps | 1 Mbps | 1 Mbps |
| Bandwidth | 10 MHz | 10 MHz | 10 MHz | 10 MHz | 10 MHz | 10 MHz | 10 MHz |
| Number of Episodes | 1000-1500 | 1000-1500 | 1000-1500 | 1000-1500 | 1000-1500 | 1000-1500 | 1500-2000 |
| DRL Action Time per Episode | 10-20 ms | 15-25 ms | 20-35 ms | 10-20 ms | 15-25 ms | 20-35 ms | 25-40 ms |
| NE Action Time per Episode | 120 ms-1 s | 150 ms-1 s | 250 ms-1 s | 120 ms-1 s | 150 ms-1 s | 250 ms-1 s | 300 ms-1 s |

The primary goal of the xApp is to improve overall network performance by maximizing throughput while ensuring fair resource allocation across users and maintaining long-term network resilience. To achieve this, the xApp dynamically adjusts user associations to balance the network load and allocates spectrum resources to maximize efficiency. While not explicitly designed for admission control, the xApp indirectly supports network stability by optimizing resource distribution, thereby preventing overload and maintaining consistent performance under varying network conditions.

Three DRL models are employed for RAN control: deep Q-network (DQN), advantage actor-critic (A2C), and multi-agent RL (MARL). DQN determines optimal actions in discrete action spaces, making it effective for single-node resource allocation. A2C combines policy optimization with value estimation for dynamic resource allocation in continuous environments. MARL addresses multi-agent coordination and interaction across multiple E2 nodes. The testbed enables analysis of the xApp's performance metrics, such as convergence rate, learning stability, and computational efficiency, under varying RAN conditions. The open-source AI models and installation instructions will be available through the OAIC repository[1].

### C. Experiments

We consider three AI models to design seven separate experiments, which are detailed in Table I. The first model, NE-DQN, is deployed as a single-agent DQN model to manage a single E2 node with 13 active UEs. The first three experiments focus on systematically varying the GA configuration across low, medium, and high setups based on specific performance thresholds derived from the indication metrics. These configurations include the generation number, population size, GA mutation rate, elitism, and crossover rate. Table I also outlines key DRL parameters such as the learning rate and gamma ($\gamma$), where $\gamma$ represents the discount factor in RL that prioritizes long-term rewards. As previously discussed, we employ adaptive tuning mechanisms for the GA parameters, ensuring they dynamically align with the system's performance and available compute resources. This adaptability is achieved through the *Indication Metric*, which is derived from the DRL agent's performance, and the *Scaling Factor*, determined by querying K8s APIs for resource availability. The *Scaling Factor*, which ranges between 0 and 1, represents the relative resource availability within the system. A value of 1 indicates maximum capacity is available for GA optimization processes, while 0 reflects minimum or no capacity. Setting the *Scaling Factor* to 1 in these experiments assumes sufficient resources are available to fully exploit the GA configurations for optimization. These adaptive mechanisms enable efficient utilization of GA settings in real time. The NE-DQN experiments provide a targeted evaluation of the NE algorithm's capacity to refine DRL xApp behavior within the computational constraints of the RAN environment.

The second model, NE-A2C, operates as a single-agent A2C model tasked with managing a single base station also handling 13 UEs. Lastly, the MARL model is deployed to manage two E2 nodes serving a total of 28 UEs. This multi-agent setup reflecting the complexity challenges of coordinating multiple agents within the same environment. Experiments four through seven, as outlined in Table I, systematically vary the GA configurations, incorporating adaptive tuning mechanisms to analyze performance and resource utilization under diverse network conditions. The NE-integrated models are particularly evaluated for their ability to balance exploration and exploitation in these increasingly complex scenarios.

By examining both single-agent and multi-agent setups, we evaluate the F-ONRL framework's scalability and its ability to maintain robust performance across varying levels of network complexity. Throughout these experiments, key network parameters such as the NE interval, maximum UE downlink traffic, operating frequency, bandwidth, transmit power, reward function, and success level are held constant across the experiments to ensure consistency and comparability.

### D. Metrics

We evaluate the proposed F-ONRL framework using a variety of metrics, including convergence rate, achieved reward, and learning stability, which captures the effectiveness of the models achieving the target performance. Also, we measure action processing time per episode (time step), a critical metric for assessing the computational overhead introduced by the integration of NE optimization into the DRL xApp.

### E. Results

Figure 4 illustrates the outcomes of the seven experiments, demonstrating the effects of integrating NE through GA into the DQN, A2C, and MARL models. Each subfigure contrasts the baseline models (GA Not Activated) with the NE-augmented models (GA Activated), offering insights into the impact of GA-triggered optimization under various configurations. In these experiments we assume the worst case scenario, where the GA triggers when the reward drops to zero.

---
[1] https://github.com/openaicellular

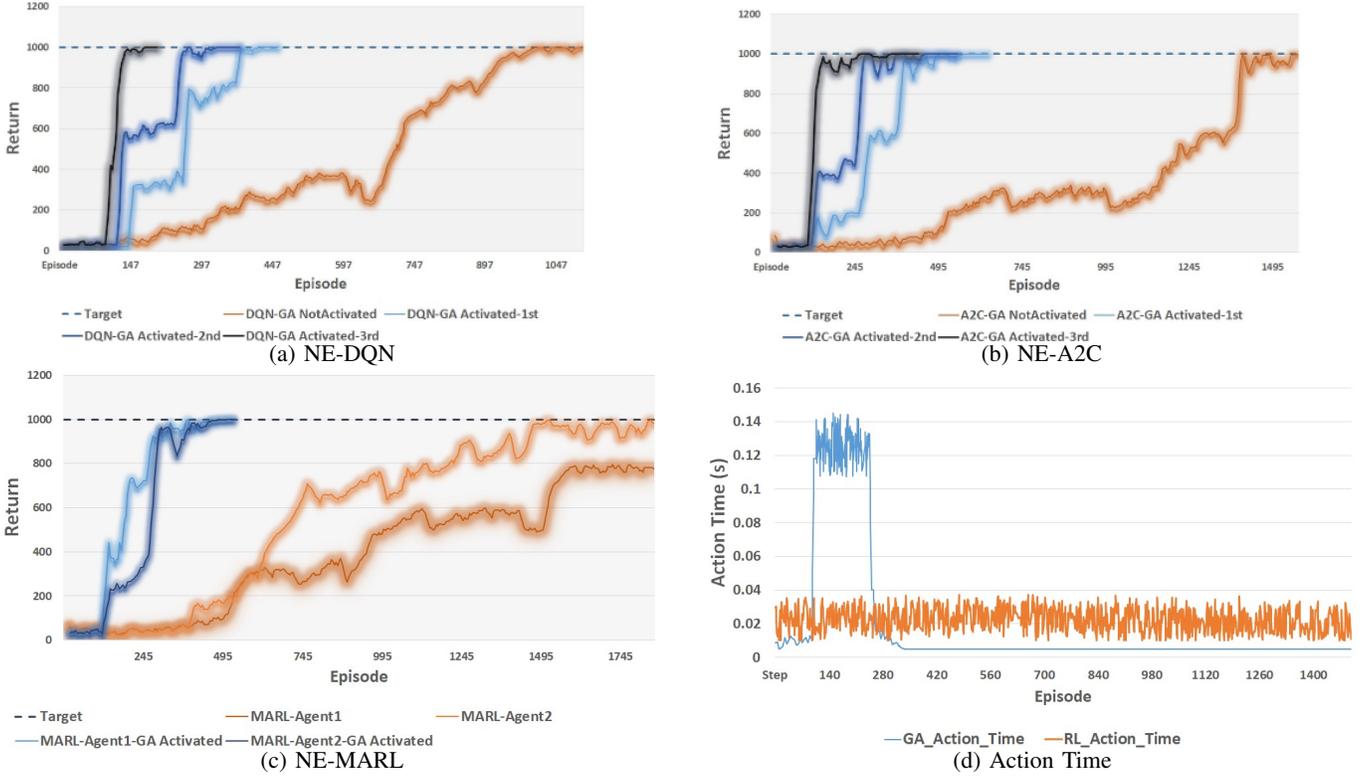

Fig. 4: Results for NE-DQN (a), NE-A2C (b), and NE-MARL (c) models, compared to their respective baselines (GA Not Activated) for the seven experiments of Table I. Actions times of NE (GA) and DRL agents (d).

In the single-agent experiments (experiments 1–3, Table I), the DQN baseline (DQN-GA Not Activated) shows slower convergence and stagnates at suboptimal reward levels due to limited exploration and susceptibility to local optima. In contrast, the DQN-GA Activated results demonstrate significantly faster convergence and higher stability. These gains are achieved through GA-driven NE optimization, which effectively fine-tunes the DNN weights to navigate the action space more effectively. The reward trajectories highlight a consistent improvement as GA configurations evolve from low to medium to high (Experiments 1, 2, and 3).

Similarly, in the A2C-based Experiments 4–6, the baseline model (A2C-GA Not Activated) exhibits slower learning and higher sensitivity to environmental dynamics, such as varying UE numbers and traffic conditions. The NE-augmented results (A2C-GA Activated) show improved learning stability and faster convergence across all configurations. By balancing exploration and exploitation more effectively, the GA fine-tuning combats hyperparameter sensitivity and enables the A2C model to achieve the target reward faster.

In the multi-agent scenario (Experiment 7), baseline agents (MARL-Agent1 and MARL-Agent2) display disparate performances, with Agent2 stalling at a reward of 800 due to coordination inefficiencies and local optima. Once GA is activated, both agents (MARL-Agent1-GA Activated and MARL-Agent2-GA Activated) reach the target reward of 1000, underscoring the ability of NE to recalibrate agent interactions and optimize multi-agent learning in complex environments.

The computational dynamics are depicted in Fig. 4(d), showing that GA operations introduce higher processing times compared to DRL. During activation episodes, GA incurs additional computational overhead because of its intensive optimization processes. However, because the F-ONRL framework deploys the NE and DRL agents in separate K8s pods with dedicated computational resources, the DRL action times remain consistent, demonstrating the benefit of F-ONRL, where GA optimizations execute in parallel without interfering with the DRL decision-making. This parallelism ensures that the framework maintains responsiveness while benefiting from NE's global optimization capabilities. Although the F-ONRL architecture necessitates additional computational resources compared to conventional DRL deployment, it shifts computation from frequent, inefficient DRL updates to less frequent but more effective NE-based exploration.

These experiments also highlight the critical role of GA parameter selection for balancing performance gains and computational costs. Fine-tuning these parameters, as outlined in Table I, enhances the learning process while minimizing resource consumption, demonstrating the adaptability of the F-ONRL framework to account for different network constraints.

*F. Experimental Challenges*

Parameter tuning poses a notable challenge during implementation, as inappropriate GA configurations could increase computational complexity and significantly impact DRL agent convergence speed and model stability. We employ a Bayesian optimization algorithm to efficiently search and identify optimal hyperparameters for GA configuration, resulting in stable convergence and reduced computational overhead. The RMR messaging library ensures timely and secure parameter exchanges between decentralized DRL agents and the centralized NE optimizer. The OAIC platform facilitates the integration of the developed xApps into the near-RT RIC interfacing with the RAN software through the E2 interface.

## IV. Research Directions

Typically, O-RAN deployments have to simultaneously optimize multiple objectives, such as spectral efficiency, service latency, and fairness, in large-scale wireless networks serving numerous end devices with heterogeneous service requests. The direct encoding capabilities of the F-ONRL xApp presented in this paper empower each genome to be structured in such a way to represent both the network topology and its specialization for a particular O-RAN metric of interest. The cooperative fitness function then evaluates each genome based on the metrics it optimizes while considering the combined performance of multiple xApps. After that, the crossover and mutation procedures create a capable offspring to address the multi-objective optimization function with adjusted weights. However, encoding all weights and connections in such large-scale deployments can lead to extensive computing overhead during the evolutionary process. We recommend two research avenues for scaling the proposed framework: First, employing transfer learning to initialize the weights of NE populations from previously optimized models can accelerate model convergence in similar O-RAN use cases with similar objectives. Secondly, multiple versions of the multi-objective F-ONRL framework can be mapped to decentralized near-RT RICs to collaboratively evolve the AI models. This can be augmented with incremental evolution where only subsets of weights/layers are evolved per cycle per decentralized RIC.

## V. Conclusion

Through an in-depth examination of the theoretical foundations and practical challenges of DRL for xApps in O-RAN's near-RT RIC, we argue for the transition toward Hybrid NE techniques. F-ONRL aligns closely with the inherent structure of DRL, offering natural and effective means of learning from scalar reward signals in complex and dynamic wireless network environments. This transition represents not merely an incremental adjustment, but a significant shift in methodology, with the potential to unlock new levels of performance and efficiency of DRL used for network operation and resource management in O-RAN. While the presented F-ONRL model incurs more computing than traditional DRL-based xApp solutions, its performance benefits overweigh the overhead. We identify targeted research for large-scale O-RAN deployments, where scalability can be achieved through offline and transfer learning combined with distributed computing which aligns with O-RAN's dual-RIC architecture and enhanced deployment solutions.

## Acknowledgement

This work was supported in part by the National Science Foundation under grant number 2120442 and by the Office of Naval Research under Award No. N00014-23-1-2808.

## References

[1] A. S. Abdalla, P. S. Upadhyaya, V. K. Shah, and V. Marojevic, "Toward next generation open radio access networks–what O-RAN can and cannot do!" *IEEE Network*, pp. 1–8, 2022.

[2] M. Kouchaki and V. Marojevic, "Actor-critic network for O-RAN resource allocation: xApp design, deployment, and analysis," in *2022 IEEE Globecom Workshops (GC Wkshps)*, 2022, pp. 968–973.

[3] M. Kouchaki *et al.*, "OpenAI dApp: An open AI platform for distributed federated reinforcement learning apps in O-RAN," in *2023 IEEE Future Networks World Forum (FNWF)*, 2023, pp. 1–6.

[4] O. Orhan *et al.*, "Connection management xApp for O-RAN RIC: A graph neural network and reinforcement learning approach," in *2021 20th IEEE International Conference on Machine Learning and Applications (ICMLA)*, 2021, pp. 936–941.

[5] I. Tamim, S. Aleyadeh, and A. Shami, "Intelligent O-RAN traffic steering for URLLC through deep reinforcement learning," *arXiv preprint arXiv:2303.01960*, 2023.

[6] S. S. Mousavi, M. Schukat, and E. Howley, "Deep reinforcement learning: an overview," in *Proceedings of SAI Intelligent Systems Conference (IntelliSys) 2016: Volume 2*. Springer, 2018, pp. 426–440.

[7] G. Dulac-Arnold, D. Mankowitz, and T. Hester, "Challenges of real-world reinforcement learning," *arXiv preprint arXiv:1904.12901*, 2019.

[8] K. Stanley *et al.*, "Designing neural networks through neuroevolution," *Nature Machine Intelligence*, vol. 1, 01 2019.

[9] P. Li, J. Thomas, X. Wang, A. Khalil, A. Ahmad, R. Inacio, S. Kapoor, A. Parekh, A. Doufexi, A. Shojaeifard, and R. J. Piechocki, "RLOps: Development life-cycle of reinforcement learning aided open RAN," *IEEE Access*, vol. 10, pp. 113 808–113 826, 2022.

[10] F. Lotfi, O. Semiari, and F. Afghah, "Evolutionary deep reinforcement learning for dynamic slice management in O-RAN," in *2022 IEEE Globecom Workshops (GC Wkshps)*, 2022, pp. 227–232.

[11] E. Galván and P. Mooney, "Neuroevolution in deep neural networks: Current trends and future challenges," *IEEE Transactions on Artificial Intelligence*, vol. 2, no. 6, pp. 476–493, 2021.

[12] M. Kouchaki, S. B. H. Natanzi, M. Zhang, B. Tang, and V. Marojevic, "O-RAN performance analyzer: Platform design, development, and deployment," *IEEE Communications Magazine*, pp. 1–8, 2024.

[13] P. S. Upadhyaya, A. S. Abdalla, V. Marojevic, J. H. Reed, and V. K. Shah, "Prototyping next-generation O-RAN research testbeds with SDRs," *arXiv preprint arXiv:2205.13178*, 2022.

## Biographies

**Mohammadreza Kouchaki** (mk1682@msstate.edu) is a PhD student in Electrical and Computer Engineering at Mississippi State University, MS, USA. His research interests focus on integrating AI and Machine Learning techniques into communication networks, with emphasis on network security, anomaly detection, resource management, and ORAN.

**Aly Sabri Abdalla** (asa298@msstate.edu) is an Assistant Research Professor in the Department of Electrical and Computer Engineering at Mississippi State University, Starkville, MS, USA. His research interests are on wireless communication and networking, software radio, spectrum sharing, wireless testbeds and testing, and wireless security.

**Vuk Marojevic** (vuk.marojevic@msstate.edu) is a professor in electrical and computer engineering at Mississippi State University, Starkville, MS, USA. His research interests include software radios, vehicle-to-everything communications, and wireless security with application to cellular communications, O-RAN, mission-critical networks, and unmanned aircraft systems.